\documentclass[sigconf]{acmart}

\AtBeginDocument{%
  }

\usepackage{multirow}       
\usepackage{enumitem}
\usepackage{balance}
\newcommand{\lsm}[1]{\textcolor{black}{{#1}}}

\newcommand{\std}[1]{{\scriptsize$\pm$#1}}
\copyrightyear{2026}
\acmYear{2026}
\setcopyright{cc}
\setcctype{by}
\acmConference[MM '26]{Proceedings of the 34th ACM International Conference on Multimedia}{November 10--14, 2026}{Rio de Janeiro, Brazil}
\acmBooktitle{Proceedings of the 34th ACM International Conference on Multimedia (MM '26), November 10--14, 2026, Rio de Janeiro, Brazil}
\acmDOI{10.1145/3767308.3836574}
\acmISBN{979-8-4007-2213-4/2026/11}
\begin{document}

\title[RankGround]{RankGround: Efficient High-Resolution GUI Grounding
via Lightweight Reranker-Guided Crop Selection}

\author{Liyang Fan}
\orcid{0009-0008-8877-6415}
\affiliation{%
  \department{College of Computer Science and Software Engineering}
  \institution{Shenzhen University}
  \city{Shenzhen}
  \country{China}
}
\additionalaffiliation{%
  \institution{Shenzhen University of Advanced Technology}
  \city{Shenzhen}
  \country{China}
}
\additionalaffiliation{%
  \department{Shenzhen Key Laboratory for High Performance Data Mining}
  \institution{Shenzhen Institutes of Advanced Technology}
  \city{Shenzhen}
  \country{China}
}
\email{ly.fan2@siat.ac.cn}

\author{Xinping Bi}
\orcid{0009-0008-0295-8200}
\affiliation{%
  \department{Shenzhen Key Laboratory for High Performance Data Mining}
  \institution{Shenzhen Institutes of Advanced Technology, Chinese Academy of Sciences}
  \city{Shenzhen}
  \country{China}
}
\email{xp.bi@siat.ac.cn}

\author{Yitai Li}
\orcid{0009-0003-7048-3837}
\affiliation{%
  \department{Shenzhen Key Laboratory for High Performance Data Mining}
  \institution{Shenzhen Institutes of Advanced Technology, Chinese Academy of Sciences}
  \city{Shenzhen}
  \country{China}
}
\email{yt.li5@siat.ac.cn}

\author{Shuaimin Li}
\authornote{Corresponding authors.}
\orcid{0000-0002-8368-916X}
\affiliation{%
  \department{Shenzhen Key Laboratory for High Performance Data Mining}
  \institution{Shenzhen Institutes of Advanced Technology, Chinese Academy of Sciences}
  \city{Shenzhen}
  \country{China}
}
\email{sm.li2@siat.ac.cn}

\author{Hui Li}
\orcid{0000-0001-9139-3855}
\affiliation{%
  \institution{Xiamen University}
  \city{Xiamen}
  \country{China}
}
\email{hui@xmu.edu.cn}

\author{Min Yang}
\authornotemark[3]
\orcid{0000-0003-3814-2728}
\affiliation{%
  \institution{Shenzhen Institutes of Advanced Technology, Chinese Academy of Sciences}
  \city{Shenzhen}
  \country{China}
}
\affiliation{%
  \institution{Shenzhen University of Advanced Technology}
  \city{Shenzhen}
  \country{China}
}
\email{min.yang@siat.ac.cn}

\renewcommand{\shortauthors}{Liyang Fan et al.}

\begin{abstract}
Graphical User Interface (GUI) grounding is a fundamental perception task for multimodal agents, enabling them to interpret natural language instructions and interact with digital interfaces. Existing methods face a fundamental trade-off between accuracy and efficiency: direct full-image inference often fails to capture small or visually similar UI elements, while multi-crop strategies improve localization at the cost of multiple expensive Vision-Language Model (VLM) calls per query.

To address this challenge, we propose \textbf{RankGround}, a two-stage framework that achieves accurate GUI grounding with a single VLM call per query. Central to our approach is \textbf{GroundRanker}, a lightweight multimodal reranker that identifies the most promising crop from a dense candidate set. Because no off-the-shelf ranking dataset is available, we construct ranking supervision data from existing grounding datasets. A strict containment criterion and boundary-aware positive augmentation improve alignment and spatial coverage in cluttered layouts. GroundRanker is then trained with a two-stage curriculum: a pointwise objective first learns coarse containment, and a listwise objective refines subtle semantic and spatial distinctions among visually similar crops.
Experimental results show that RankGround consistently outperforms strong baselines while reducing computational cost. It achieves \textbf{1.4× faster inference} and improves localization accuracy by \textbf{5.5\%} on average over the second-best method across all backbones and screen scales, establishing a new state of the art in both efficiency and precision for GUI grounding.
\end{abstract}



\maketitle

\section{Introduction}

\begin{figure}[t]
  \centering
  \includegraphics[width=\linewidth]{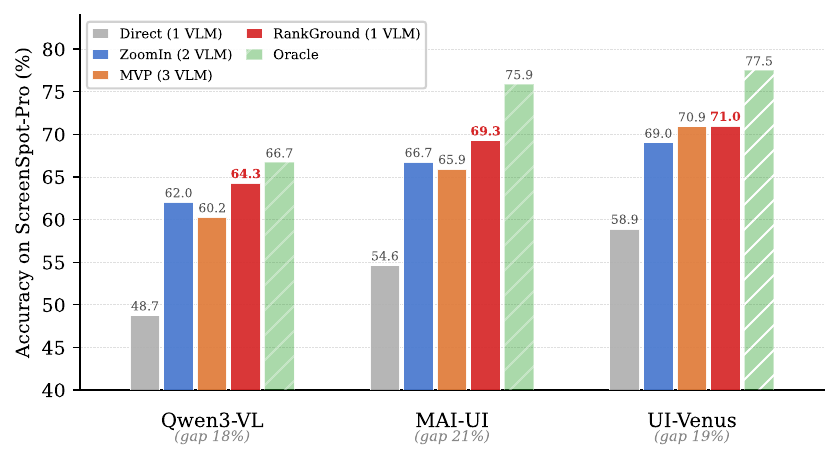}
  \caption{Accuracy of GUI grounding methods on ScreenSpot-Pro (8B scale). The gap between Direct inference and the Oracle upper bound (over 18\,\%) reveals room for improvement. Multi-crop methods (ZoomIn, MVP) close this gap but require two to three VLM calls per query. RankGround matches or exceeds their accuracy with a single VLM call.}
  \Description{Five bars are shown for each backbone. For Qwen3-VL, the accuracies from Direct through Oracle are 48.7, 62.0, 60.2, 64.3, and 66.7 percent. For MAI-UI, they are 54.6, 66.7, 65.9, 69.3, and 75.9 percent. For UI-Venus, they are 58.9, 69.0, 70.9, 71.0, and 77.5 percent. RankGround is the strongest non-Oracle method for every backbone.}
  \label{fig:motivation}
\end{figure}

GUI agents automate tasks across digital platforms~\cite{cheng2024seeclick,qin2025ui,zhou2025maiui}. Their success depends on \emph{GUI grounding}, which maps a natural language instruction to a location on the screen~\cite{gou2025navigating}. A correct task plan still fails when the agent cannot locate the target element~\cite{zhou2025maiui,zhang2025mvp}. Higher display resolutions and denser layouts make small or visually similar elements harder to distinguish~\cite{li2025screenspotpro}.

Existing GUI grounding methods typically use \textbf{direct inference} or \textbf{multiple crops}~\cite{zhou2025maiui}.
\textbf{Direct inference} sends the entire screenshot through a vision-language model (VLM) once to predict target coordinates~\cite{cheng2024seeclick,gou2025navigating,gao2026ui,zhou2025maiui}. It is fast, but small elements lose detail when a high-resolution screenshot is resized for the model.
\textbf{Multi-crop} methods partition the screen and apply the VLM to each region~\cite{zhang2025mvp,zhou2025maiui}. The crops expose finer detail, but the VLM now runs repeatedly. ZoomIn~\cite{zhou2025maiui} uses two calls per query, whereas MVP~\cite{zhang2025mvp} often uses three or more. Latency and memory use grow with the crop count, which complicates real-time deployment. Figure~\ref{fig:motivation} shows the resulting cost. Similar elements in different crops can also lead to conflicting predictions.

\textbf{RankGround} separates region selection from coordinate prediction. Overlap tiling or sliding windows produce a dense candidate set that covers potential target locations. A lightweight multimodal reranker, \textbf{GroundRanker}, scores these crops, and the GUI-specialist VLM predicts coordinates from the top-ranked crop. The VLM therefore runs once per query while retaining local detail.

Crop selection remains difficult in dense layouts because several regions may contain similar elements. We convert instance annotations from existing grounding datasets into crop-level supervision. A crop is positive only when it fully encloses the target under a \emph{strict containment criterion}. \emph{Boundary-aware positive augmentation} then places the target near crop boundaries to reduce position bias.

GroundRanker follows a two-stage curriculum. Pointwise binary cross-entropy first teaches containment, and a listwise objective then separates near-miss crops from valid ones. We fine-tune Qwen3-VL-Reranker-2B~\cite{li2026qwen3vlreranker} with \emph{Low-Rank Adaptation (LoRA)}~\cite{HuSWALWWC22} on the query and value projections in its vision and language components. This setup learns GUI-specific spatial alignment while preserving pretrained multimodal features.

RankGround turns crop selection into a small ranking problem and reserves the costly VLM for coordinate prediction.
Our main contributions follow.

\begin{itemize}
    \item \textbf{RankGround} decouples region selection from coordinate prediction and uses one GUI-specialist VLM call per query.
    \item \textbf{GroundRanker} learns crop selection from strict containment labels, boundary-aware augmentation, and a pointwise-to-listwise curriculum.
    \item Across all tested backbones and scales, RankGround improves average localization accuracy by \textbf{5.5\%} over the second-best method and runs \textbf{1.4× faster}.
\end{itemize}

\section{Related Work}

\subsection{GUI Grounding}

Early GUI grounding work adapted visual grounding methods developed for natural images~\cite{plummer2015flickr30k, mao2016generation, qiao2021referring, kamath2021mdetr, li2022glip, yang2022unitab}. Some models add hierarchical low-rank adaptation~\cite{xiao2024hivg,deng2021transvg} or multimodal weight modulation~\cite{yao2024mmca, li2022glip}. Such methods transfer poorly to GUI screenshots, where elements are structured and often occupy only a few pixels after resizing~\cite{cheng2024seeclick, li2025screenspotpro}. Their scale distribution also differs sharply from natural-image datasets such as RefCOCO~\cite{yu2016modeling}.

GUI-specific models now use dedicated data and pre-training. SeeClick~\cite{cheng2024seeclick} introduced ScreenSpot and showed the value of GUI pre-training. UGround~\cite{gou2025navigating}, UI-TARS~\cite{qin2025ui}, MAI-UI~\cite{zhou2025maiui}, and UI-Venus-1.5~\cite{gao2026ui} scale this approach with more data or larger models. ScreenSpot-Pro~\cite{li2025screenspotpro} nevertheless shows that small targets on high-resolution interfaces remain difficult.

Recent methods recover visual detail through extra inference stages. ZoomIn~\cite{zhou2025maiui} uses coarse-to-fine refinement. ScreenSeekeR~\cite{li2025screenspotpro} and ZoomClick~\cite{jiang2025zoomclick} repeatedly zoom into a predicted region. GOLD~\cite{lee2025gold} proposes regions at low resolution and revisits them at high resolution. MVP~\cite{zhang2025mvp} aggregates predictions from attention-guided views, whereas CoG~\cite{li2025cog} refines coordinates through self-reasoning.
Training-based methods learn when or how to add these stages. GUI-ARP~\cite{ye2025guiarp} combines supervised fine-tuning with GRPO so that the model can request another inference stage. Focus~\cite{tang2025focus} learns to switch between fast prediction and a slower reasoning path. Both still spend extra forward passes on selection or verification. RankGround assigns region selection to a lightweight reranker and calls the grounding VLM once.

\subsection{Multimodal Reranking}

Reranking is common in information retrieval. A fast first stage narrows the candidate set before a cross-encoder scores the remaining items~\cite{nogueira2019reranking}. Recent systems extend this design to multimodal retrieval~\cite{wei2024uniir, faysse2025colpali, lin2025mmrerank}. Qwen3-VL-Reranker~\cite{li2026qwen3vlreranker} uses a Qwen3-VL cross-encoder to score text-image query-document pairs.
RankGround applies multimodal reranking to GUI grounding and uses crop relevance scores for region selection. To our knowledge, this is the first use of a multimodal reranker for this task.


\begin{figure*}[t]
  \centering
  \includegraphics[width=0.95\textwidth]{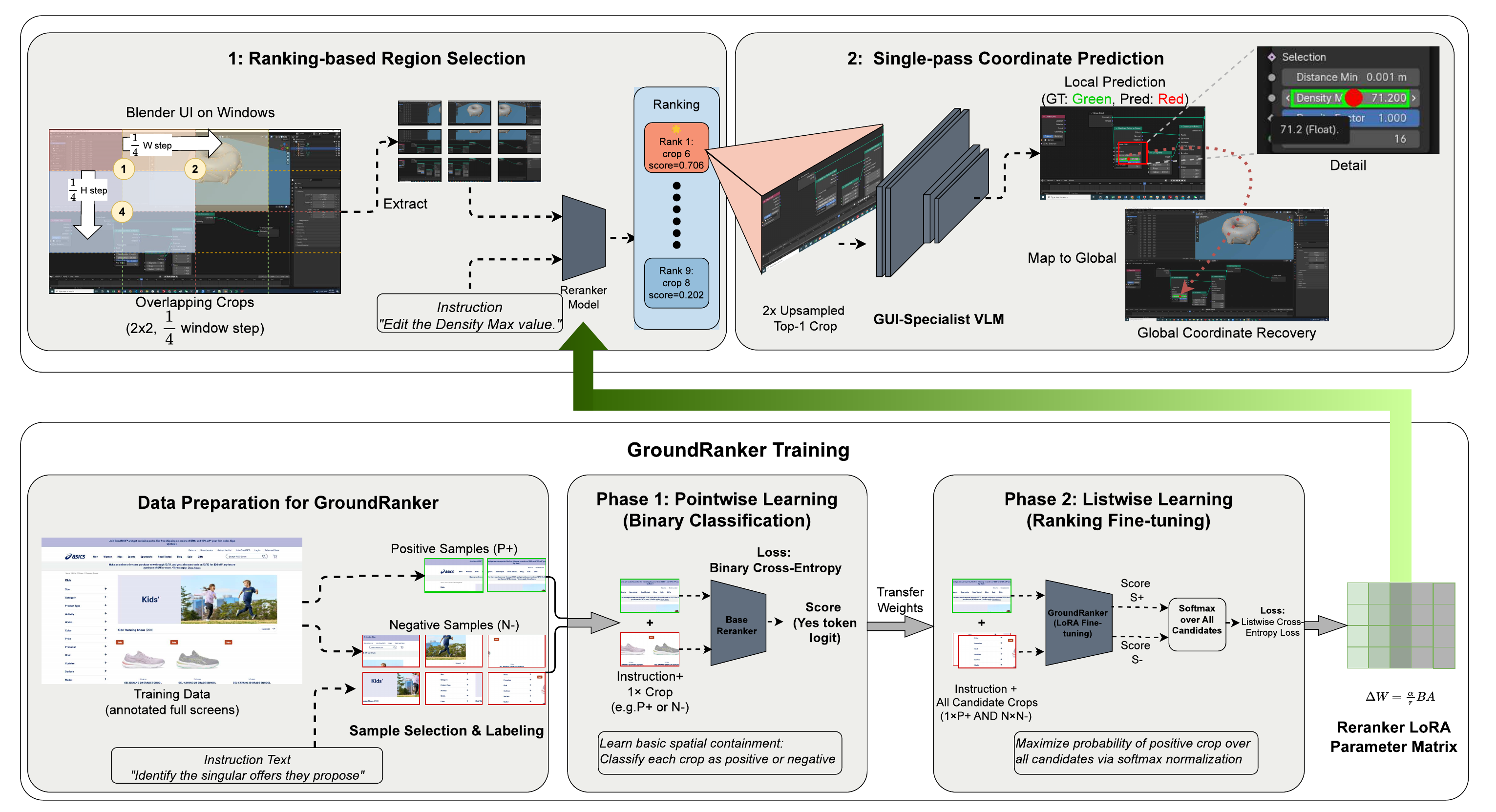}
  \caption{Overview of the RankGround pipeline. Given a screenshot and a natural language instruction, RankGround partitions the image into $N$ crops via an overlap tiling strategy, ranks them with a fine-tuned multimodal reranker, and invokes a GUI-specialist VLM \emph{once} on the selected crop to predict coordinates, which are mapped back to the original image space.}

  \Description{The upper half traces inference from a Blender screenshot and instruction through nine overlapping crops. The reranker scores all crops and selects crop 6 with score 0.706. That crop is enlarged twofold, processed by the GUI-specialist VLM, and mapped back to the full screenshot. The lower half traces training. Annotated screenshots yield positive and negative crops, pointwise binary classification teaches containment, and listwise fine-tuning compares all candidates while updating LoRA parameters.}
  \label{fig:architecture}
\end{figure*}

\section{Method}
\subsection{Overview}
\label{sec:overview}

Given a screenshot $\mathbf{I} \in \mathbb{R}^{H \times W \times 3}$ and a natural language instruction $q$,
GUI grounding aims to predict the normalized click coordinate $\hat{\mathbf{p}} \in [0,1]^2$ of the target UI element on the screenshot.

Existing multi-crop approaches ask the same VLM to select a region and predict its coordinates. The VLM runs on every candidate crop, which is costly at high resolutions.

Region selection does not require precise coordinates. It only needs to identify the crop that contains the target. We therefore treat selection as a lightweight \emph{ranking} problem and leave one high-resolution coordinate prediction to the VLM.

Let $\mathcal{C} = \{c_1, \ldots, c_N\}$ be the candidate set. Each region $c_k$ has a top-left offset $\mathbf{o}_k$ and spatial extent $\mathbf{d}$. Prediction follows
\begin{align}
  c^{*} &= \arg\max_{c_k \in \mathcal{C}}\; s(q, c_k), \\
  \hat{\mathbf{p}} &= \mathbf{o}^{*} + F_\phi(q, c^{*}) \odot \mathbf{d},
\end{align}
where $s(\cdot)$ ranks the candidate regions and $F_\phi$ is a GUI-specialist VLM applied once to the selected region. We instantiate $s$ as \textbf{GroundRanker}. Section~\ref{sec:reranker} describes how it scores crops from a coverage-preserving tiling. Section~\ref{sec:coord} gives the coordinate mapping for the selected crop. Figure~\ref{fig:architecture} shows the full framework.


\subsection{Ranking-based Region Selection}
\label{sec:reranker}
Ranking-based selection narrows the downstream VLM input to one crop. Section~\ref{sec:partition} defines candidate generation, and \lsm{Section~\ref{sec:data_preparation}} derives containment labels and augmented positives. \lsm{Section~\ref{sec:reranker_training}} trains GroundRanker on these crops. At inference, \lsm{Section~\ref{sec:reranker_inference}} passes the top crop to coordinate prediction.
\subsubsection{Overlap Tiling-based Candidate Generation}
\label{sec:partition}

A candidate must contain the full target. Otherwise, no reranker can recover it, bounding end-to-end accuracy.

Formally, let $B^{*} = [x_1^{*}, y_1^{*}, x_2^{*}, y_2^{*}]$ denote the ground-truth bounding box of the target element. 
We say that a candidate set $\mathcal{C}$ achieves \emph{full coverage} if and only if
\begin{equation}
  \exists\; c_k \in \mathcal{C} \;\;\text{s.t.}\;\; B^{*} \subseteq \mathrm{box}(c_k),
  \label{eq:coverage}
\end{equation}
where $\mathrm{box}(c_k)$ denotes the spatial extent of crop $c_k$. 
Violating this condition makes errors irrecoverable, regardless of modeling.

We use \textbf{Overlap Tiling}, a $G{\times}G$ grid with overlap ratio $\rho$, to satisfy this condition. 
Each crop spans $\lfloor H/G \rfloor \times \lfloor W/G \rfloor$, and adjacent crops are shifted by $(1-\rho)$ of their extent along both axes, resulting in $N = (2G{-}1)^2$ candidates in total. 
For targets smaller than a crop, $\rho = 0.5$ guarantees Eq.~\eqref{eq:coverage} by applying the interval covering lemma along each axis. The crop at position $(i,j)$ starts at
\begin{equation*}
  x_{ij} = \left\lfloor \frac{i}{2(G-1)} W \right\rfloor, \quad 
  y_{ij} = \left\lfloor \frac{j}{2(G-1)} H \right\rfloor,
\end{equation*}
with crop width $w_c = \lceil W/G \rceil$ and height $h_c = \lceil H/G \rceil$, for $i,j \in \{0, \ldots, 2G-2\}$.
Each crop is resized to the VLM input resolution.

The size assumption matches nearly all ScreenSpot-Pro samples. Under the default $G{=}2$ layout, each crop covers about 25\% of the screenshot and overlaps its neighbors. The resulting Oracle coverage is 99.75\%. The 90th-percentile target occupies only 0.156\% of the screenshot for text and 0.034\% for icons. A wide target may cross every crop boundary, so Eq.~\eqref{eq:coverage} no longer guarantees full containment. GroundRanker can still choose a useful partial view. Figure~\ref{fig:wide_target} shows a ScreenSpot-V2 example in which the target spans crop boundaries. The selected crop scores 0.90, compared with 0.39 for the runner-up, and the VLM predicts the correct click.


\subsubsection{Data Preparation for GroundRanker}
~\label{sec:data_preparation}
Existing grounding datasets lack crop-level ranking labels. We derive them automatically from screenshots with ground-truth element annotations.

For a screenshot $\mathbf{I}$ of size $W \times H$, sliding-window tiling scales the window $(w,h)$ to a fixed pixel budget $N_{\max}$ according to
\begin{equation}
r = \sqrt{\frac{N_{\max}}{W \cdot H}}, \quad w = \lfloor r \cdot W \rfloor, \quad h = \lfloor r \cdot H \rfloor.
\end{equation}
The window moves in steps of $(w/2,h/2)$ for 50\% overlap and snaps to the image boundary to avoid incomplete edge crops. Images within the pixel budget ($W \cdot H \leq N_{\max}$) use the $2 \times 2$ overlapping grid from Section~\ref{sec:partition}.

Each crop $c_k$ receives a binary \textbf{containment} label. It is positive only when it fully contains the target box $B^{*} = [x_1^{*}, y_1^{*}, x_2^{*}, y_2^{*}]$ under the condition
\begin{equation}
  \lsm{l_k \leq x_1^{*} ;\wedge\; t_k \leq y_1^{*} ;\wedge\; l_k{+}w \geq x_2^{*} ;\wedge\; t_k{+}h \geq y_2^{*},}
\end{equation} where $(l_k, t_k)$ denotes the top-left corner of the $k$-th candidate crop,
ensuring that the target element is entirely visible. This strict criterion is stronger than center-based or IoU-based labeling and aligns directly with the coverage condition~\eqref{eq:coverage}.

\textbf{Boundary-aware positive augmentation} counters the center bias of uniform windows. Let $\Delta_x = w - (x_2^{*} - x_1^{*})$ and $\Delta_y = h - (y_2^{*} - y_1^{*})$ denote the slack between the crop and target box. Each positive crop yields up to 12 variants. We place the target at the four corners and four edge midpoints, then draw four more positions near the boundary. The resulting samples spread positives across the crop.

Each \textbf{listwise~\cite{cao2007listwise}} sample pairs one positive crop or augmented variant with every negative crop from the same image and the instruction $q$. This matches inference and exposes the full contrast among candidates to the loss in Eq.~\eqref{eq:listwise}.

\subsubsection{GroundRanker Training}
~\label{sec:reranker_training}
GroundRanker is a vision-language cross-encoder that scores how likely each crop is to contain the target. It selects
\begin{equation}
  c^{*} = c_{k^{*}}, \quad k^{*} = \arg\max_k\; s_\theta(q, c_k).
  \label{eq:select}
\end{equation}

where $s_\theta$ jointly scores the crop and instruction $q$.

The cross-encoder processes $(q,c)$ jointly to retain fine spatial correspondence. We use the affirmative token logit as its ranking score,
\begin{equation}
  s_\theta(q, c) = \mathrm{logit}_{\texttt{yes}}\bigl(f_\theta(q, c)\bigr),
  \label{eq:score}
\end{equation}
where $f_\theta$ denotes the reranker. The score is comparable across crops and needs no classification head.

GroundRanker uses a \textbf{pointwise-to-listwise curriculum}.

\textit{Stage~1 (pointwise).} Binary cross-entropy teaches spatial containment,
\begin{equation}
  \mathcal{L}_1 = \mathbb{E}_{(q,\,c,\,y)}\bigl[-y \log \sigma(s_\theta) - (1{-}y) \log(1 - \sigma(s_\theta))\bigr],
  \label{eq:bce}
\end{equation}
where $y \in {0,1}$. Boundary-aware positives reduce position bias.

\textit{Stage~2 (listwise).} A listwise loss contrasts valid crops with near-miss crops that overlap the target but fail containment,
\begin{equation}
  \mathcal{L}_2 = -\sum_{c \in \mathcal{C}} \tilde{y}_c \cdot
                \log \frac{\exp\bigl(s_\theta(q,c)\bigr)}{\sum_{c' \in \mathcal{C}} \exp\bigl(s_\theta(q,c')\bigr)},
  \label{eq:listwise}
\end{equation}
where $\tilde{y}_c = y_c / \sum_{c'} y_{c'}$ normalizes labels. Stage~1 learns a coarse ranking. Stage~2 sharpens near-miss boundaries.

We adapt the pretrained cross-encoder with Low-Rank Adaptation (LoRA)~\cite{HuSWALWWC22}. Trainable low-rank matrices $A$ and $B$ update each frozen weight matrix $W_0$ as
\begin{equation}
h = W_0 x + \Delta W x = W_0 x + \frac{\alpha}{r} B A x,
\label{eq:lora}
\end{equation}
Here $x$ is the input, $h$ is the output, and $\Delta W$ is the update. The rank satisfies $r \ll \min(d,k)$, and $\alpha$ controls its scale. GroundRanker applies LoRA to query and value projections in the vision encoder's self-attention and the language model's cross-attention. This preserves pretrained features while learning spatial alignment for GUI crops.

\subsubsection{Region Selection.}~\label{sec:reranker_inference}
At inference, overlap tiling or sliding windows generate $\mathcal{C}$, and GroundRanker scores each crop. We select
\begin{equation}
  c^{*} = \arg\max_{c_k \in \mathcal{C}} s_\theta(q, c_k),
  \label{eq:inference_select}
\end{equation}
The selected crop then passes to coordinate prediction, so the VLM processes one region rather than the full candidate set.

\subsection{Coordinate Prediction and Mapping}
\label{sec:coord}

The VLM $F_\phi$ predicts a local coordinate $\hat{\mathbf{p}}' \in [0,1]^2$ within the upsampled crop $c^{*}$. A linear transform maps it back to the original image without approximation error.

Upsampling the crop by $\alpha$ yields $\alpha/G$ resolution relative to the full image.
For $\alpha \geq G$, the VLM sees more detail per element, improving small-element localization.

\textit{Complexity analysis.}
Let $T_R$ denote one batched reranker pass and $T_V$ one VLM pass. GroundRanker scores all $N$ crops in one batch, so region selection costs $\mathcal{O}(T_R)$ rather than $\mathcal{O}(N T_R)$. Coordinate prediction adds $\mathcal{O}(T_V)$, giving $\mathcal{O}(T_R + T_V)$ total latency. In our measurements, $T_R \approx 145$\,ms and $T_V$ ranges from about 210 to 810\,ms. RankGround is slower than Direct but avoids the repeated VLM calls used by ZoomIn and MVP. Local crop processing preserves detail without the $\mathcal{O}(N T_V)$ cost of running the VLM on every candidate.

\section{Experimental Setup}
\label{sec:experiments}


\textit{Dataset.}
We evaluate on \textbf{ScreenSpot-Pro}~\cite{li2025screenspotpro}, comprising
1,581 tasks from 23 professional applications across five industries and three
operating systems.
We also test cross-benchmark generalization on \textbf{UI-Vision}~\cite{nayak2025uivision}.
Table~\ref{tab:bounding_box_stats} summarizes the bounding-box size distribution for both element types.

\begin{table}[t]
\centering
\footnotesize
\caption{Bounding-box statistics for Text and Icon elements in ScreenSpot-Pro. \textbf{Rel.\ area} is the percentage of screenshot area. P10 and P90 are the 10th and 90th percentiles.}
\setlength{\tabcolsep}{2pt}
\begin{tabular*}{\columnwidth}{@{\extracolsep{\fill}}lrrrrrr@{}}
\toprule
 & \multicolumn{3}{c}{\textbf{Text} ($n{=}977$)} & \multicolumn{3}{c}{\textbf{Icon} ($n{=}604$)} \\
\cmidrule(r){2-4} \cmidrule(l){5-7}
 & \textbf{Mean} & \textbf{Med.} & \textbf{P10/P90} & \textbf{Mean} & \textbf{Med.} & \textbf{P10/P90} \\
\midrule
Width (px) & 141 & 111 & 36 / 272 & 37 & 26 & 16 / 54 \\
Height (px) & 34 & 25 & 18 / 61 & 29 & 24 & 17 / 48 \\
Area (px\textsuperscript{2}) & 4,655 & 3,186 & 828 / 7,680 & 1,764 & 624 & 305 / 2,344 \\
Rel.\ area (\%) & 0.086 & 0.065 & 0.017 / 0.156 & 0.031 & 0.014 & 0.005 / 0.034 \\
\bottomrule
\end{tabular*}

\label{tab:bounding_box_stats}
\end{table}

\textit{Evaluation Metrics.}
\textbf{Click Accuracy} counts predictions inside the target box and is micro-averaged over 977 \textbf{Text} and 604 \textbf{Icon/Widget} targets. We also report each type. Reranker ablations use \textbf{Hit Rate@$k$}, the fraction of queries with a valid crop among the top-$k$ results for the corresponding grounding instruction.

\textit{Baselines.}
Direct inference sends the full screenshot through each backbone once. ZoomIn~\cite{zhou2025maiui} predicts a coarse location and refines it in a second call. MVP~\cite{zhang2025mvp} selects attention-guided regions and aggregates several crop predictions. We test Qwen3-VL, MAI-UI, and UI-Venus-1.5 at both 8B and 2B scales. Qwen3-VL~\cite{bai2025qwen3vl} uses the basic prompt, and the other models use their native prompts.

\textit{Implementation.}
GroundRanker is trained on 6,789 ViSurf instances. Data construction yields about 295K pairwise, 124K pointwise, and 78K listwise samples. LoRA uses rank 16, scale 32, and dropout 0.05 on query and value projections. We use AdamW with a $1{\times}10^{-4}$ learning rate, 0.01 weight decay, effective batch size 8, 5\% linear warmup, and cosine decay. Training converges within one epoch on RTX 4090 GPUs. Latency is wall-clock time per sample over ScreenSpot-Pro at batch size 1, including preprocessing and post-processing but excluding data loading.

\section{Results and Analysis}
\subsection{Main Results}
\label{sec:main}

Table~\ref{tab:main} shows that RankGround leads all six backbone-scale settings with one GUI-specialist VLM call and one reranker pass per query.
\#VLM counts GUI-specialist VLM calls, and $\Delta$ is the absolute accuracy gain (\%) over Direct for the same backbone.
Bold and underlining mark each scale's best and runner-up.

\begin{table*}[t]
\centering
\caption{Comparison of GUI grounding methods on ScreenSpot-Pro at 8B and 2B scales. 
}
\label{tab:main}
\footnotesize
\setlength{\tabcolsep}{3pt}
\begin{tabular*}{\textwidth}{@{\extracolsep{\fill}}llc ccccc ccccc@{}}
\toprule
\multicolumn{3}{c}{} & \multicolumn{5}{c}{\textbf{8B Scale}} & \multicolumn{5}{c}{\textbf{2B Scale}} \\
\cmidrule(lr){4-8} \cmidrule(lr){9-13}
\textbf{Method} & \textbf{Model} & \textbf{\#VLM} & \textbf{Time (ms)} & \textbf{Acc (\%)} & \textbf{Text (\%)} & \textbf{Icon (\%)} & $\boldsymbol{\Delta}$ & \textbf{Time (ms)} & \textbf{Acc (\%)} & \textbf{Text (\%)} & \textbf{Icon (\%)} & $\boldsymbol{\Delta}$ \\
\midrule
\multirow{3}{*}{Direct}
& Qwen3-VL~\cite{bai2025qwen3vl} & \multirow{3}{*}{1} & 814\std{11} & 48.72\std{0.04} & 65.06\std{0.12} & 22.30\std{0.10} & & 642\std{2} & 35.42\std{0.06} & 48.00\std{0.10} & 15.07\std{0.17} & \\
& MAI-UI~\cite{zhou2025maiui}   & & 468\std{3} & 54.59\std{0.04} & 70.93\std{0.06} & 28.15\std{0.12} & & 235\std{3} & 43.64\std{0.04} & 58.75\std{0.08} & 19.20\std{0.10} & \\
& UI-Venus~\cite{gao2026ui} & & 480\std{23} & 58.89\std{0.06} & 74.10\std{0.08} & 34.27\std{0.15} & & 212\std{6} & 48.13\std{0.04} & 61.51\std{0.06} & 26.49\std{0.12} & \\
\addlinespace
\multirow{3}{*}{ZoomIn~\cite{zhou2025maiui}}
& Qwen3-VL~\cite{bai2025qwen3vl} & \multirow{3}{*}{2} & 1486\std{39} & 62.00\std{0.10} & 80.08\std{0.12} & 32.73\std{0.10} & +13.28 & 1130\std{7} & 47.82\std{0.08} & 62.95\std{0.12} & 23.34\std{0.19} & +12.40 \\
& MAI-UI~\cite{zhou2025maiui}   & & 981\std{12} & 66.73\std{0.11} & 81.30\std{0.47} & 43.16\std{0.47} & +12.14 & 505\std{10} & 57.37\std{0.06} & 73.29\std{0.10} & 31.62\std{0.15} & +13.73 \\
& UI-Venus~\cite{gao2026ui} & & 1008\std{3} & 69.00\std{0.11} & 82.77\std{0.12} & 46.74\std{0.48} & +10.11 & 473\std{3} & 59.20\std{0.08} & 71.24\std{0.10} & 39.74\std{0.17} & +11.07 \\
\addlinespace
\multirow{3}{*}{MVP~\cite{zhang2025mvp}}
& Qwen3-VL~\cite{bai2025qwen3vl} & \multirow{3}{*}{3} & 4648\std{56} & 60.24\std{0.04} & 78.37\std{0.06} & 30.91\std{0.19} & +11.52 & 3058\std{38} & 38.48\std{0.04} & 51.25\std{0.12} & 17.83\std{0.10} & +3.06 \\
& MAI-UI~\cite{zhou2025maiui}   & & 3500\std{48} & 65.91\std{0.06} & 80.28\std{0.06} & 42.66\std{0.10} & +11.32 & 2342\std{32} & 57.39\std{0.04} & 73.40\std{0.10} & 31.70\std{0.12} & +13.75 \\
& UI-Venus~\cite{gao2026ui} & & 4049\std{45} & \underline{70.90\std{0.04}} & \underline{83.62\std{0.06}} & \textbf{50.33\std{0.10}} & +12.01 & 2418\std{28} & 56.67\std{0.06} & 66.53\std{0.08} & \underline{40.73\std{0.12}} & +8.54 \\
\addlinespace
\multirow{3}{*}{RankGround}
& Qwen3-VL~\cite{bai2025qwen3vl} & \multirow{3}{*}{1} & 959\std{12} & 64.26\std{0.08} & 81.99\std{0.10} & 35.60\std{0.19} & +15.54 & 787\std{5} & 52.75\std{0.06} & 69.40\std{0.12} & 25.83\std{0.15} & +17.33 \\
& MAI-UI~\cite{zhou2025maiui}   & & 613\std{5} & 69.26\std{0.06} & 82.29\std{0.08} & 48.18\std{0.12} & +14.67 & 380\std{5} & \underline{63.00\std{0.04}} & \textbf{78.71\std{0.10}} & 37.58\std{0.17} & +19.36 \\
& UI-Venus~\cite{gao2026ui} & & 625\std{23} & \textbf{70.97\std{0.04}} & \textbf{84.54\std{0.06}} & \underline{49.01\std{0.10}} & +12.08 & 357\std{7} & \textbf{64.71\std{0.06}} & \underline{76.25\std{0.08}} & \textbf{46.03\std{0.15}} & +16.58 \\
\bottomrule
\end{tabular*}
\end{table*}

RankGround improves on Direct in all six configurations, with absolute gains from 12.08 to 19.36 points. These gains require one GUI-specialist VLM call and a 145\,ms reranker pass. ZoomIn uses two VLM calls. RankGround improves its average accuracy by 2.3 points at 8B and 5.4 points at 2B, and is faster for every backbone. The larger 2B gain shows that crop selection matters more when the grounder has less capacity.

RankGround also matches or exceeds MVP on every backbone. Average accuracy rises from 65.7\% to 68.2\% at 8B and from 50.9\% to 60.2\% at 2B. MVP takes 2,342--4,648\,ms per sample, or $4$--$12\times$ longer, because it evaluates several crops with the VLM.

Recent best-reported ScreenSpot-Pro results provide wider context, though they use different backbones and training protocols. ScreenSeekeR~\cite{li2025screenspotpro} combines GPT-4o with OS-Atlas-7B and reports 48.1\% after multiple calls. GUI-ARP~\cite{ye2025guiarp} reports 60.8\% with a 7B model trained by SFT and reinforcement learning. CoG~\cite{li2025cog} reaches 68.4\% with a three-call Qwen3-VL-235B/32B pipeline. RankGround reaches 64.7\% at 2B and 71.0\% at 8B with one grounding-VLM call plus one reranker pass. Table~\ref{tab:main} remains the controlled comparison. These published numbers instead show where RankGround sits among systems with different model and data budgets.

\textit{Peak memory.}
Co-resident reranker and VLM peak at 21.2\,GB for 8B and 8.6\,GB for 2B, both within 24\,GB. Sequential execution lowers them to 16.94 and 4.38\,GB, matching Direct and staying below MVP's 17.21 and 4.61\,GB.

\subsection{Ablation Studies}
\label{sec:ablation}

Section~\ref{sec:abl_partition} examines partition geometry, Section~\ref{sec:abl_reranker} examines reranker training, and Section~\ref{sec:abl_prompt} tests prompt wording. Unless stated otherwise, the ablations use UI-Venus-1.5-2B with Overlap $2{\times}2$ and report mean\,$\pm$\,std over three runs.

\subsubsection{Partitioning Strategy}
\label{sec:abl_partition}

Table~\ref{tab:partition} compares uniform, overlapping, and sliding-window grids using the base reranker. \textbf{Oracle} assumes correct crop selection, and \textbf{Hit@1} is the fraction of valid top-ranked crops. \textbf{Gain@1} subtracts random selection, where Rand@1 $= \bar{n}_+/N$, $\bar{n}_+$ is the mean positive-crop count, and $N$ is the total crop count.

\begin{table*}[t]
\centering
\caption{
  Ablation of image partitioning strategies on ScreenSpot-Pro. \textbf{Bold} marks the best result per column, and \underline{underlining} marks the second best. Time is average per-batch reranker latency in ms. Hit@5 is unavailable for Uniform $2{\times}2$ because it generates only four crops.
}
\label{tab:partition}
\footnotesize
\begin{tabular*}{\textwidth}{@{\extracolsep{\fill}}l *{8}{c} @{}}
\toprule
\multirow{2}{*}{\textbf{Layout}} & \multicolumn{2}{c}{\textbf{Coverage}}
 & \multicolumn{3}{c}{\textbf{Hit Rate@$\boldsymbol{k}$ (\%) $\boldsymbol{\uparrow}$}}
 & \multicolumn{2}{c}{\textbf{Selection Quality}}
 & \multirow{2}{*}{\textbf{Time (ms) $\boldsymbol{\downarrow}$}} \\
\cmidrule(lr){2-3} \cmidrule(lr){4-6} \cmidrule(lr){7-8}
 & \textbf{\#Crops} & \textbf{Oracle (\%) $\boldsymbol{\uparrow}$} & \textbf{@1} & \textbf{@3} & \textbf{@5} & \textbf{Rand@1 (\%)} & \textbf{Gain@1 (\%) $\boldsymbol{\uparrow}$} & \\
\midrule

Uniform $2{\times}2$ & 4 & 93.04
  & 62.66\std{0.10} & 88.49\std{0.06} & N/A
  & 23.26 & 39.40\std{0.10} & 156\std{5} \\
\quad + opt.\ prompt & 4 & 93.04
  & \textit{67.09\std{0.18}} & \textit{89.42\std{0.04}} & \textit{N/A}
  & 23.26 & \textit{43.83\std{0.18}} & 159\std{6} \\

\addlinespace
Uniform $3{\times}3$ & 9 & 90.70
  & 54.97\std{0.17} & 74.26\std{0.17} & 83.24\std{0.11}
  & 10.08 & 44.89\std{0.17} & 160\std{6} \\
\quad + opt.\ prompt & 9 & 90.70
  & \textit{61.44\std{0.26}} & \textit{79.38\std{0.06}} & \textit{86.32\std{0.07}}
  & 10.08 & \textit{51.36\std{0.26}} & 163\std{5} \\

\addlinespace
Uniform $4{\times}4$ & 16 & 85.90
  & 45.86\std{0.29} & 62.98\std{0.42} & 70.15\std{0.19}
  & 5.37 & 40.49\std{0.29} & 200\std{8} \\
\quad + opt.\ prompt & 16 & 85.90
  & \textit{51.53\std{0.46}} & \textit{67.95\std{0.35}} & \textit{75.35\std{0.07}}
  & 5.37 & \textit{46.16\std{0.46}} & 196\std{7} \\

\midrule
Overlap $2{\times}2$ & 9 & 99.75
  & 67.13\std{0.13} & 88.32\std{0.10} & 94.73\std{0.04}
  & 23.84 & 43.29\std{0.13} & 145\std{4} \\
\quad + opt.\ prompt & 9 & 99.75
  & \textit{72.51\std{0.35}} & \textit{91.21\std{0.17}} & \textit{96.16\std{0.04}}
  & 23.84 & \textit{48.67\std{0.35}} & 148\std{5} \\

\addlinespace
Overlap $3{\times}3$ & 25 & 99.43
  & 56.27\std{0.42} & 75.29\std{0.13} & 83.49\std{0.33}
  & 9.50 & 46.78\std{0.42} & 280\std{12} \\
\quad + opt.\ prompt & 25 & 99.43
  & \textit{63.23\std{0.19}} & \textit{81.07\std{0.44}} & \textit{87.92\std{0.46}}
  & 9.50 & \textit{53.73\std{0.19}} & 275\std{11} \\

\midrule
Sliding 1080p & $\sim$7.2 & 99.75
  & 77.25\std{0.04} & 93.28\std{0.13} & 92.86\std{0.15}
  & 42.44 & 34.81\std{0.04} & 155\std{6} \\
\quad + opt.\ prompt & $\sim$7.2 & 99.75
  & \textit{80.71\std{0.33}} & \textit{93.64\std{0.04}} & \textit{93.69\std{0.13}}
  & 42.44 & \textit{38.27\std{0.33}} & 158\std{7} \\

\addlinespace
Sliding 1080sq & $\sim$14.8 & 99.43
  & 65.72\std{0.17} & 84.06\std{0.17} & 91.08\std{0.16}
  & 21.96 & 43.75\std{0.17} & 190\std{8} \\
\quad + opt.\ prompt & $\sim$14.8 & 99.43
  & \textit{71.54\std{0.33}} & \textit{87.73\std{0.25}} & \textit{93.21\std{0.17}}
  & 21.96 & \textit{49.57\std{0.33}} & 194\std{9} \\

\addlinespace
Sliding 720p & $\sim$17.7 & 99.62
  & 64.54\std{0.26} & 81.72\std{0.11} & 89.05\std{0.13}
  & 17.79 & 46.75\std{0.26} & 210\std{9} \\
\quad + opt.\ prompt & $\sim$17.7 & 99.62
  & \textit{70.59\std{0.32}} & \textit{86.27\std{0.13}} & \textit{91.49\std{0.17}}
  & 17.79 & \textit{52.80\std{0.32}} & 207\std{8} \\

\addlinespace
Sliding 720sq & $\sim$35.2 & 98.80
  & 54.33\std{0.35} & 73.77\std{0.29} & 80.83\std{0.22}
  & 8.70 & 45.63\std{0.35} & 380\std{15} \\
\quad + opt.\ prompt & $\sim$35.2 & 98.80
  & \textit{60.57\std{0.10}} & \textit{78.37\std{0.29}} & \textit{84.69\std{0.23}}
  & 8.70 & \textit{51.87\std{0.10}} & 386\std{14} \\

\midrule
LoRA FT reranker, Overlap $2{\times}2$ & 9 & 99.75 & \underline{85.52\std{0.19}} & \underline{93.36\std{0.11}} & \underline{97.41\std{0.08}} & 23.84 & \underline{61.68\std{0.19}} & 145\std{4} \\
\quad + opt.\ prompt & 9 & 99.75 & \textbf{88.17\std{0.14}} & \textbf{94.69\std{0.08}} & \textbf{97.72\std{0.06}} & 23.84 & \textbf{64.33\std{0.14}} & 147\std{5} \\

\bottomrule
\end{tabular*}
\end{table*}

\textit{Oracle coverage.}
Finer uniform grids cover less of the screen per crop and split more boundary elements. Their Oracle accuracy falls from 93.0\% at $2{\times}2$ to 85.9\% at $4{\times}4$. Overlapping grids and sliding windows instead keep Oracle above 98.8\%, so a 50\% overlap contains nearly every target in at least one crop.

\textit{Reranker discriminability.}
Overlap $2{\times}2$ reaches 43.3\,\% Gain@1, or 48.7\,\% with the optimized prompt, while retaining 99.75\,\% Oracle. Sliding 1080p has the highest raw Hit@1 at 77.3\,\%, rising to 80.7\,\% with the prompt. Its random baseline is already 42.4\,\% because it produces only 7.2 crops on average, leaving 34.8\,\% Gain@1.
Finer sliding windows (720p, 720sq) improve Gain@1 but degrade Hit@1 as the
reranker faces more candidates with similar visual content.
Overlap $3{\times}3$ reaches 46.8\,\% Gain@1, or 53.7\,\% with the prompt, but uses 25 crops rather than 9.

\textit{Efficiency.}
Overlap $2{\times}2$ scores 9 crops in 145\,ms. Overlap $3{\times}3$ takes 280\,ms for 25 crops, and sliding windows take up to 380\,ms for 35 crops. Batched reranker latency therefore grows roughly with the crop count.

\begin{table}[!b]
\centering
\caption{Default and optimized reranker prompts.}
\label{tab:prompt_compare}
\small
\begin{tabular*}{\columnwidth}{@{\extracolsep{\fill}}lp{0.72\columnwidth}@{}}
\hline
\textbf{Default}   & \textit{``Given a search query, retrieve relevant
                     candidates that answer the query.''} \\[4pt]
\textbf{Optimized} & \textit{``Given a GUI operation instruction, determine
                     whether the provided image region contains the target UI
                     element described in the instruction.''} \\
\hline
\end{tabular*}
\end{table}

\begin{figure}[t]
  \centering
  \includegraphics[width=\linewidth]{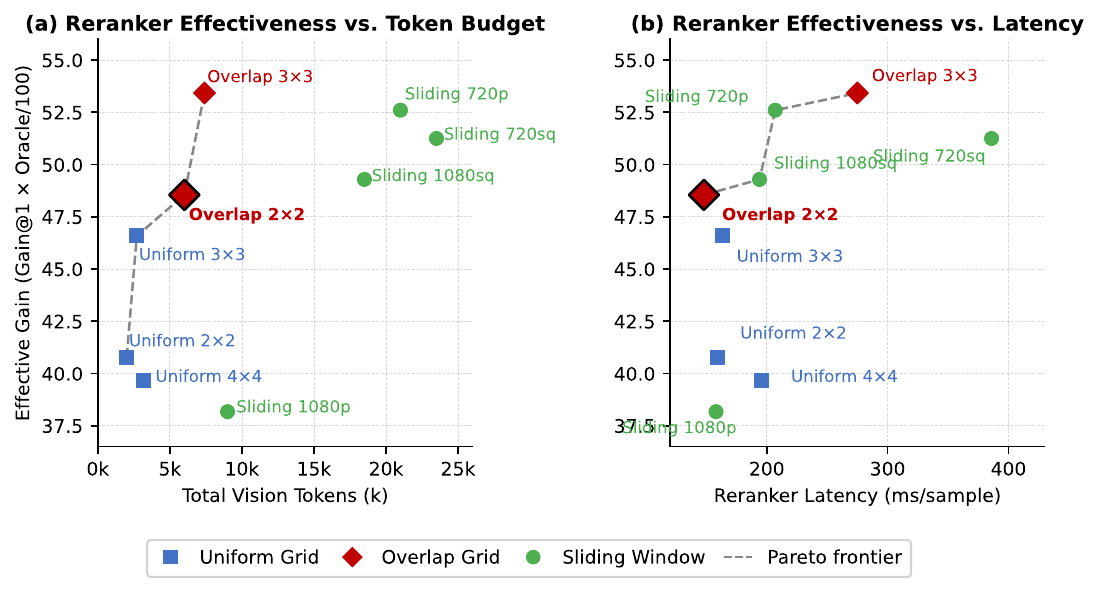}
  \caption{Effective Gain ($\text{Gain@1} \times \text{Oracle}$) vs.\ vision token budget (left) and reranker latency (right) across strategies. }
  \Description{Uniform grids appear as blue squares, overlapping grids as red diamonds, and sliding windows as green circles. Overlap 2x2 lies on the dashed Pareto frontier in both panels. Overlap 3x3 has the largest effective gain but uses more tokens and nearly twice the latency. Sliding 720p approaches that gain with more than three times the token budget, while the uniform grids have lower gain.}
  \label{fig:strategy_selection}
\end{figure}

Overlap $2{\times}2$ reaches 99.75\,\% Oracle with a 9-crop search space and 43.3\,\% Gain@1. Its 145\,ms latency is lower than larger layouts. We use it as RankGround's default.

\subsubsection{Reranker Design}
\label{sec:abl_reranker}

Region selection quality, not VLM capacity, limits multi-crop grounding. With $2{\times}$ crops, the Oracle upper bound exceeds every multi-crop method by more than 7 points across the 8B backbones. Figure~\ref{fig:motivation} shows this gap.
Higher crop resolution further benefits downstream grounding even under
perfect region selection, with $2{\times}$ outperforming $1{\times}$ under
Oracle conditions ($+$3.67\,\% for Qwen3-VL, $+$1.98\,\% for MAI-UI,
$+$2.91\,\% for UI-Venus).
The base reranker remains more than 10 points below Oracle accuracy, which motivates the LoRA fine-tuning in Section~\ref{sec:reranker}. Across three seeds, the pointwise-to-listwise curriculum reaches 88.2\,\% Hit@1 and improves 2B Text accuracy by 0.62 points over pointwise-only training. Section~\ref{sec:curriculum_analysis} reports the full comparison.

\subsubsection{Prompt Design}
\label{sec:abl_prompt}

The chat template stores the task definition in \texttt{<Instruct>}, the grounding instruction in \texttt{<Query>}, and the crop in \texttt{<Document>}.
Table~\ref{tab:prompt_compare} contrasts the \textbf{default prompt} inherited
from Qwen3-VL-Reranker with our \textbf{optimized prompt}, which asks directly
whether the crop contains the target.
The optimized prompt consistently improves Hit@1 over the default prompt across
all partitioning strategies. The largest gain is $+$5.4 points on Overlap
$2{\times}2$, shown by the italicized rows of Table~\ref{tab:partition}.
The prompt gives the reranker a clearer containment question without changing its weights.

\subsection{Reranker Generalization}
\label{sec:generalization}

Table~\ref{tab:hr_metrics} shows 78.70\,\% Hit@1 on \textbf{UI-Vision}~\cite{nayak2025uivision} without benchmark-specific fine-tuning.

\begin{table}[!htbp]
\centering
\caption{
  Hit Rate (HR) on out-of-distribution benchmarks using Overlap $2{\times}2$ and UI-Venus-1.5-2B.
}
\label{tab:hr_metrics}
\footnotesize
\begin{tabular*}{\columnwidth}{@{\extracolsep{\fill}}llccc@{}}
\toprule
\textbf{Method} & \textbf{Benchmark} & \textbf{HR@1 (\%)} & \textbf{HR@3 (\%)} & \textbf{HR@5 (\%)} \\
\midrule
Base reranker          & UI-Vision     & 52.93\std{0.11} & 79.02\std{0.09} & 90.16\std{0.08} \\
\addlinespace[0.5ex]
RankGround             & UI-Vision     & 78.70\std{0.00} & 89.05\std{0.00} & 95.26\std{0.00} \\
\bottomrule
\end{tabular*}
\end{table}

Table~\ref{tab:oracle_acc} reports 49.00\,\% end-to-end accuracy for RankGround on UI-Vision, compared with 18.88\,\% for Direct.

\begin{table}[!htbp]
\centering
\caption{
  Oracle coverage and Accuracy on out-of-distribution benchmarks using Overlap $2{\times}2$ and Qwen3-VL-2B.
}
\label{tab:oracle_acc}
\footnotesize
\begin{tabular*}{\columnwidth}{@{\extracolsep{\fill}}llcc@{}}
\toprule
\textbf{Method} & \textbf{Benchmark} & \textbf{Oracle (\%)} & \textbf{Acc (\%)} \\
\midrule
Direct                 & UI-Vision     & N/A  & 18.88\std{0.06} \\
\addlinespace[0.5ex]
Base reranker          & UI-Vision     & 98.36 & 25.76\std{0.10} \\
\addlinespace[0.5ex]
RankGround             & UI-Vision     & 98.36 & 49.00\std{0.06} \\ 
\bottomrule
\end{tabular*}
\end{table}

The reranker is trained only on ViSurf~\cite{liu2025visurf}, an OmniACT-derived dataset with 79\% desktop and 21\% web data from 62 applications. ScreenSpot-Pro and UI-Vision therefore differ from its training distribution. Table~\ref{tab:uivision_bbox_stats} shows that UI-Vision also differs from ScreenSpot-Pro. It contains 2,786 Icon samples and 758 Text samples. Element sizes are similar in pixels, but their relative areas are about $2{\times}$ larger, which indicates lower-resolution screenshots on average.

\begin{table}[!htbp]
\centering
\footnotesize
\setlength{\tabcolsep}{2pt}
\caption{Bounding-box statistics for Text and Icon elements in UI-Vision. \textbf{Rel.\ area} is the percentage of screenshot area. P10 and P90 are the 10th and 90th percentiles.}
\begin{tabular*}{\columnwidth}{@{\extracolsep{\fill}}lrrrrrr@{}}
\toprule
 & \multicolumn{3}{c}{\textbf{Text} ($n{=}758$)} & \multicolumn{3}{c}{\textbf{Icon} ($n{=}2{,}786$)} \\
\cmidrule(r){2-4} \cmidrule(l){5-7}
 & \textbf{Mean} & \textbf{Med.} & \textbf{P10/P90} & \textbf{Mean} & \textbf{Med.} & \textbf{P10/P90} \\
\midrule
Width (px) & 136 & 108 & 38 / 275 & 41 & 34 & 22 / 58 \\
Height (px) & 32 & 26 & 19 / 51 & 32 & 28 & 20 / 46 \\
Area (px\textsuperscript{2}) & 4,498 & 3,002 & 870 / 10,135 & 1,435 & 962 & 459 / 2,592 \\
Rel.\ area (\%) & 0.209 & 0.161 & 0.045 / 0.454 & 0.069 & 0.052 & 0.027 / 0.112 \\
\bottomrule
\end{tabular*}
\label{tab:uivision_bbox_stats}
\end{table}

\subsection{Broader Benchmarks}
\label{sec:broader_benchmarks}

We further test UI-Venus-1.5-8B on five benchmarks that cover web, desktop, and operating-system interfaces. Table~\ref{tab:broader_benchmarks} uses the same grounder for Direct, ZoomIn, and RankGround within every benchmark.

\begin{table}[!htbp]
\centering
\caption{Click accuracy (\%) on five additional benchmarks with UI-Venus-1.5-8B. $\Delta$ is the gain over ZoomIn.}
\label{tab:broader_benchmarks}
\footnotesize
\setlength{\tabcolsep}{2pt}
\begin{tabular*}{\columnwidth}{@{\extracolsep{\fill}}lrrrr@{}}
\toprule
\textbf{Benchmark} & \textbf{Direct} & \textbf{ZoomIn} & \textbf{RankGround} & $\boldsymbol{\Delta}$ \\
\midrule
ScreenSpot-V2     & \textbf{96.31} & 94.18 & 95.36 & $+1.18$ \\
MMBench-GUI-L2    & 87.48 & 85.84 & \textbf{89.84} & $+4.00$ \\
OSWorld-G         & 74.71 & 72.55 & \textbf{78.04} & $+5.49$ \\
OSWorld-G-Refined & 80.20 & 80.59 & \textbf{84.71} & $+4.12$ \\
VenusBench-GD     & 84.71 & 85.54 & \textbf{89.62} & $+4.08$ \\
\bottomrule
\end{tabular*}
\end{table}

RankGround beats ZoomIn on all five benchmarks by 1.18--5.49 points. It also exceeds Direct by 2.36--4.91 points on MMBench-GUI-L2, OSWorld-G, OSWorld-G-Refined, and VenusBench-GD. ScreenSpot-V2 is the one exception: Direct is already at 96.31\%, while RankGround stays within one point and remains above ZoomIn. The gains on the other four sets show that learned crop selection transfers beyond ScreenSpot-Pro and UI-Vision.

\subsection{Curriculum Analysis}
\label{sec:curriculum_analysis}

We compare the two-stage curriculum with Pointwise-only and Listwise-only training. The data, LoRA configuration, optimizer, and evaluation protocol remain fixed. Each result is averaged over three random seeds. Pointwise-only uses $\mathcal{L}_1$ throughout, while Listwise-only starts directly from $\mathcal{L}_2$ without containment pretraining.

\begin{table}[!htbp]
\centering
\caption{Reranker accuracy (\%) for each training objective on ScreenSpot-Pro. Results are mean$\pm$std over three runs.}
\label{tab:curriculum_ranking}
\footnotesize
\setlength{\tabcolsep}{2pt}
\begin{tabular*}{\columnwidth}{@{\extracolsep{\fill}}lrrrrr@{}}
\toprule
\textbf{Objective} & \textbf{Hit@1} & \textbf{Hit@3} & \textbf{Hit@5} & \textbf{Text@1} & \textbf{Icon@1} \\
\midrule
Pointwise-only & 87.86\std{0.18} & 94.24\std{0.11} & 97.91\std{0.07} & 92.84\std{0.22} & \textbf{79.80\std{0.31}} \\
Listwise-only & 88.17\std{0.16} & 94.75\std{0.10} & 97.98\std{0.06} & 93.86\std{0.19} & 78.97\std{0.34} \\
Two-Stage & \textbf{88.20\std{0.15}} & \textbf{94.82\std{0.09}} & \textbf{98.05\std{0.06}} & \textbf{94.28\std{0.18}} & 78.36\std{0.30} \\
\bottomrule
\end{tabular*}
\end{table}

Table~\ref{tab:curriculum_ranking} shows that the curriculum gives the best Hit@1, Hit@3, Hit@5, and Text Hit@1. Pointwise-only remains strongest on icons. Small icons often need only a binary containment decision, while the listwise stage helps more when several text-bearing crops are semantically similar. The curriculum also reaches 90.00\% validation AUROC and completes its listwise stage at step 260, 30 steps earlier than Listwise-only.

\begin{table}[!htbp]
\centering
\caption{End-to-end accuracy (\%) for each reranker objective on ScreenSpot-Pro. Results are mean$\pm$std over three runs.}
\label{tab:curriculum_e2e}
\footnotesize
\setlength{\tabcolsep}{2pt}
\begin{tabular*}{\columnwidth}{@{\extracolsep{\fill}}llrrr@{}}
\toprule
\textbf{Objective} & \textbf{Model} & \textbf{Acc} & \textbf{Text} & \textbf{Icon} \\
\midrule
Pointwise-only & Qwen3-VL-2B & 52.62\std{0.32} & 68.58\std{0.41} & \textbf{26.82\std{0.58}} \\
Listwise-only & Qwen3-VL-2B & 52.75\std{0.30} & 68.78\std{0.39} & \textbf{26.82\std{0.56}} \\
Two-Stage & Qwen3-VL-2B & \textbf{52.90\std{0.28}} & \textbf{69.20\std{0.35}} & 26.53\std{0.54} \\
\midrule
Pointwise-only & Qwen3-VL-8B & 65.72\std{0.28} & 83.01\std{0.33} & 37.75\std{0.51} \\
Listwise-only & Qwen3-VL-8B & 65.34\std{0.30} & 82.50\std{0.35} & 37.58\std{0.53} \\
Two-Stage & Qwen3-VL-8B & \textbf{65.85\std{0.26}} & \textbf{83.20\std{0.31}} & \textbf{37.78\std{0.49}} \\
\bottomrule
\end{tabular*}
\end{table}

The ranking gains carry into coordinate prediction. Two-Stage gives the highest overall and Text accuracy at both model scales. At 2B, it improves overall accuracy by 0.28 points over Pointwise-only and Text accuracy by 0.62 points. The smaller gains at 8B suggest that a stronger coordinate predictor can absorb some crop-selection noise, but it still benefits from the curriculum.

\begin{figure}[!htbp]
  \centering
  \includegraphics[width=\linewidth]{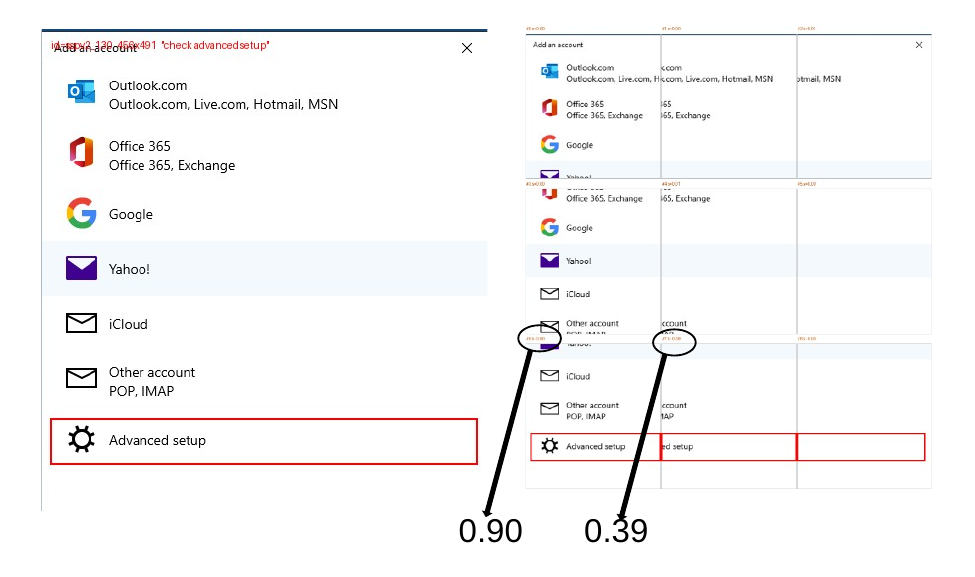}
  \Description{The left panel shows an email account setup window. A red outline marks the wide Advanced setup row near the bottom. The right panel divides the same window into nine overlapping crops. The target spans boundaries across the bottom row. Arrows identify a crop containing the useful portion of the target with score 0.90 and a competing crop with score 0.39.}
  \caption{Wide-target example on ScreenSpot-V2. The target crosses crop boundaries, yet GroundRanker selects the useful crop with a clear score margin.}
  \label{fig:wide_target}
\end{figure}

\subsection{Failure Modes and Fallbacks}
\label{sec:failures}

Overlap Tiling covers the target in 99.75\% of ScreenSpot-Pro samples, while GroundRanker reaches 88.2\% Hit@1. Coverage failures therefore account for only 0.25\% of the data. Most first-stage errors come from choosing the wrong candidate among valid crops. They concentrate on pure icons: the reranker error rate is 19.9\% for icons and 5.5\% for text. Repeated visual motifs and weak textual cues make icons harder to distinguish. Without the target in the selected crop, the coordinate predictor cannot recover.

The errors also cluster by application. Table~\ref{tab:error_by_app} lists the ten hardest applications in ScreenSpot-Pro. Origin has a 41.9\% error rate, almost twice the 21.0\% rate of Windows Common. Its toolbars contain many chart-type icons with similar shapes, and instructions such as ``plot candlestick chart'' provide little local text for matching.

\begin{table}[!htbp]
\centering
\caption{Applications with the highest GroundRanker error rates on ScreenSpot-Pro.}
\label{tab:error_by_app}
\footnotesize
\setlength{\tabcolsep}{2pt}
\begin{tabular*}{\columnwidth}{@{\extracolsep{\fill}}lrrrr@{}}
\toprule
\textbf{Application} & \textbf{OS} & \textbf{Total} & \textbf{Errors} & \textbf{Err. (\%)} \\
\midrule
Origin          & Win & 62 & 26 & \textbf{41.9} \\
Windows Common  & Win & 81 & 17 & 21.0 \\
Android Studio  & mac & 80 & 14 & 17.5 \\
DaVinci Resolve & mac & 44 & 7  & 15.9 \\
macOS Common    & mac & 65 & 10 & 15.4 \\
Premiere Pro    & Win & 52 & 8  & 15.4 \\
Fruitloops      & Win & 57 & 8  & 14.0 \\
PyCharm         & mac & 78 & 9  & 11.5 \\
Vivado          & Win & 80 & 9  & 11.2 \\
MATLAB          & mac & 93 & 9  & 9.7 \\
\bottomrule
\end{tabular*}
\end{table}

\begin{figure}[!htbp]
  \centering
  \includegraphics[width=\linewidth]{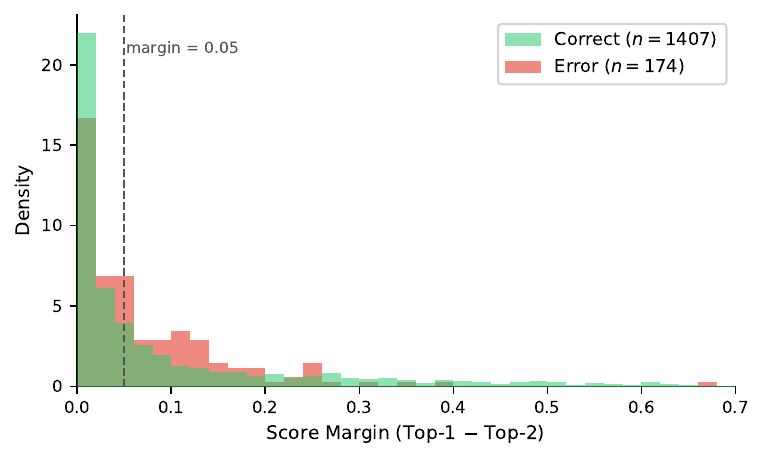}
  \Description{The horizontal axis spans score margins from zero to 0.7 and the vertical axis shows density. Correct selections, drawn in green, include 1,407 samples. Errors, drawn in salmon, include 174 samples. Both distributions peak near zero, but errors are concentrated below 0.2. Correct cases have a longer sparse tail reaching about 0.65. A vertical dashed line marks 0.05.}
  \caption{Score margins for correct and incorrect crop selections. The dashed line marks a margin of 0.05, below which 54.6\% of errors fall.}
  \label{fig:margin_hist}
\end{figure}

The score margin gives a second view of these errors. Among incorrect selections, 54.6\% have a Top-1 minus Top-2 margin below 0.05, while only 7.5\% exceed 0.2. Most errors are therefore close decisions between plausible crops rather than confident mismatches. Figure~\ref{fig:margin_hist} shows this concentration near zero.

\begin{figure}[!htbp]
  \centering
  \includegraphics[width=\linewidth]{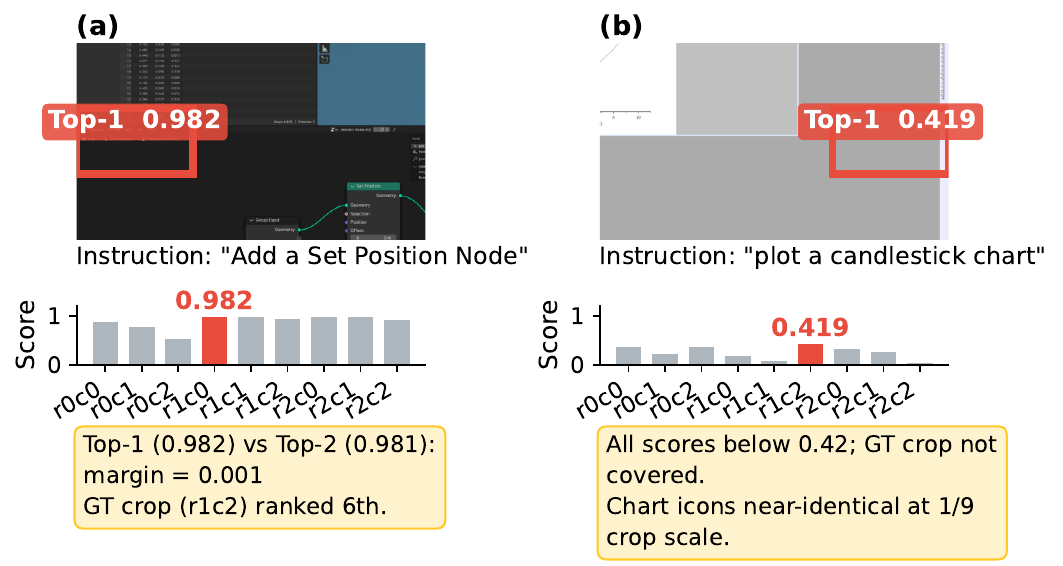}
  \Description{Panel a uses the Blender instruction Add a Set Position Node. The selected crop scores 0.982, the runner-up scores 0.981, and the ground-truth crop is ranked sixth. Panel b uses the Origin instruction plot a candlestick chart. The selected crop scores 0.419, every candidate remains below 0.42, and the selected region does not cover the ground-truth chart icon.}
  \caption{Representative failures on ScreenSpot-Pro. The red box marks the selected crop. Panel (a) is a near tie, while panel (b) contains visually similar chart icons.}
  \label{fig:failure_cases}
\end{figure}

Figure~\ref{fig:failure_cases} shows the two main failure patterns. In the Blender case, several crops contain plausible matches and the top two scores differ by only 0.001. In the Origin case, the chart icons remain hard to tell apart at crop scale, and every score stays below 0.42.

We also test whether extra crops can rescue these cases. For Top-$k$, the grounding VLM processes each of the highest-ranked crops, then coordinate-token confidence chooses the final prediction. Top-1 gives 71.0\% accuracy with 1.0 VLM call per query. Top-2 gives 64.0\% with 2.0 calls, and Top-3 gives 56.9\% with 3.0 calls. A confidence-triggered fallback reaches 68.4\% with 1.47 calls on average. Coordinate-token confidence does not compare predictions across crops as reliably as GroundRanker compares the crops themselves. We therefore retain Top-1, which gives the best measured accuracy and the lowest cost.

\section{Limitations}

RankGround depends on one selected crop. Its 99.75\% Oracle coverage assumes that the target is smaller than a crop, so wide controls such as Figure~\ref{fig:wide_target} have no formal containment guarantee. Hit@1 is 88.2\%, and our Top-$k$ experiments show that coordinate-token confidence cannot reliably recover a wrong selection. A better fallback would need confidence calibrated across crops.

GroundRanker was trained on ViSurf, whose data are 79\% desktop and 21\% web. Transfer to UI-Vision and five other benchmarks is encouraging, but mobile-heavy, multilingual, and changing interfaces remain untested. Pure icons also have a 19.9\% selection error rate versus 5.5\% for text. The fixed 145\,ms reranker pass is modest beside an 8B VLM but more visible at 2B. RankGround is faster than repeated-call baselines, not Direct, and offers an accuracy--efficiency trade-off.

\section{Conclusion}

On ScreenSpot-Pro, the gap between Direct and Oracle exceeds 18 points. Region selection, rather than coordinate prediction, accounts for most of this gap. An Oracle selector with one VLM call also beats ZoomIn and MVP by more than 7 points across the 8B backbones.

RankGround replaces repeated VLM-based selection with GroundRanker. Strict containment labels and boundary-aware augmentation teach it which crops preserve a target, while the pointwise-to-listwise curriculum separates near misses. Across backbones and scales, it gains 5.5\,\% over the best multi-crop baseline and runs 1.4$\times$ faster.

\clearpage
\begin{acks}
This work is supported by the New Generation Artificial Intelligence-National
Science and Technology Major Project (No.~2025ZD0122702), the Natural Science
Foundation of Guangdong Province of China (2024A1515030166,
2025B1515020032), and the Innovation Team Project of Guangdong Province
(No.~2024KCXTD017).
\end{acks}

\bibliographystyle{ACM-Reference-Format}
\balance
\bibliography{main}

@inproceedings{cheng2024seeclick,
  author       = {Kanzhi Cheng and
                  Qiushi Sun and
                  Yougang Chu and
                  Fangzhi Xu and
                  Yantao Li and
                  Jianbing Zhang and
                  Zhiyong Wu},
  editor       = {Lun{-}Wei Ku and
                  Andre Martins and
                  Vivek Srikumar},
  title        = {SeeClick: Harnessing {GUI} Grounding for Advanced Visual {GUI} Agents},
  booktitle    = {Proceedings of the 62nd Annual Meeting of the Association for Computational
                  Linguistics (Volume 1: Long Papers), {ACL} 2024, Bangkok, Thailand,
                  August 11-16, 2024},
  pages        = {9313--9332},
  publisher    = {Association for Computational Linguistics},
  year         = {2024},
  doi          = {10.18653/V1/2024.ACL-LONG.505},
  bibsource    = {dblp computer science bibliography, https://dblp.org}
}

@inproceedings{gou2025navigating,
  author       = {Boyu Gou and
                  Ruohan Wang and
                  Boyuan Zheng and
                  Yanan Xie and
                  Cheng Chang and
                  Yiheng Shu and
                  Huan Sun and
                  Yu Su},
  title        = {Navigating the Digital World as Humans Do: Universal Visual Grounding
                  for {GUI} Agents},
  booktitle    = {The Thirteenth International Conference on Learning Representations,
                  {ICLR} 2025, Singapore, April 24-28, 2025},
  publisher    = {OpenReview.net},
  year         = {2025},
  url          = {https://openreview.net/forum?id=kxnoqaisCT},
}

@article{qin2025ui,
  author       = {Yujia Qin and
                  Yining Ye and
                  Junjie Fang and
                  Haoming Wang and
                  Shihao Liang and
                  Shizuo Tian and
                  Junda Zhang and
                  Jiahao Li and
                  Yunxin Li and
                  Shijue Huang and
                  Wanjun Zhong and
                  Kuanye Li and
                  Jiale Yang and
                  Yu Miao and
                  Woyu Lin and
                  Longxiang Liu and
                  Xu Jiang and
                  Qianli Ma and
                  Jingyu Li and
                  Xiaojun Xiao and
                  Kai Cai and
                  Chuang Li and
                  Yaowei Zheng and
                  Chaolin Jin and
                  Chen Li and
                  Xiao Zhou and
                  Minchao Wang and
                  Haoli Chen and
                  Zhaojian Li and
                  Haihua Yang and
                  Haifeng Liu and
                  Feng Lin and
                  Tao Peng and
                  Xin Liu and
                  Guang Shi},
  title        = {{UI-TARS:} Pioneering Automated {GUI} Interaction with Native Agents},
  journal      = {CoRR},
  volume       = {abs/2501.12326},
  year         = {2025},
  url          = {https://doi.org/10.48550/arXiv.2501.12326},
  doi          = {10.48550/ARXIV.2501.12326},
  eprinttype   = {arXiv},
  eprint       = {2501.12326},
  bibsource    = {dblp computer science bibliography, https://dblp.org}
}

@inproceedings{li2025screenspotpro,
  author       = {Kaixin Li and
                  Ziyang Meng and
                  Hongzhan Lin and
                  Ziyang Luo and
                  Yuchen Tian and
                  Jing Ma and
                  Zhiyong Huang and
                  Tat{-}Seng Chua},
  editor       = {Cathal Gurrin and
                  Klaus Schoeffmann and
                  Min Zhang and
                  Luca Rossetto and
                  Stevan Rudinac and
                  Duc{-}Tien Dang{-}Nguyen and
                  Wen{-}Huang Cheng and
                  Phoebe Chen and
                  Jenny Benois{-}Pineau},
  title        = {ScreenSpot-Pro: {GUI} Grounding for Professional High-Resolution Computer
                  Use},
  booktitle    = {Proceedings of the 33rd {ACM} International Conference on Multimedia,
                  {MM} 2025, Dublin, Ireland, October 27-31, 2025},
  pages        = {8778--8786},
  publisher    = {{ACM}},
  year         = {2025},
  doi          = {10.1145/3746027.3755688},
  bibsource    = {dblp computer science bibliography, https://dblp.org}
}

@article{zhou2025maiui,
  author       = {Hanzhang Zhou and
                  Xu Zhang and
                  Panrong Tong and
                  Jianan Zhang and
                  Liangyu Chen and
                  Quyu Kong and
                  Chenglin Cai and
                  Chen Liu and
                  Yue Wang and
                  Jingren Zhou and
                  Steven Hoi},
  title        = {{MAI-UI} Technical Report: Real-World Centric Foundation {GUI} Agents},
  journal      = {CoRR},
  volume       = {abs/2512.22047},
  year         = {2025},
  url          = {https://doi.org/10.48550/arXiv.2512.22047},
  doi          = {10.48550/ARXIV.2512.22047},
  eprinttype   = {arXiv},
  eprint       = {2512.22047},
  bibsource    = {dblp computer science bibliography, https://dblp.org}
}

@article{gao2026ui,
  author       = {Venus Team and
                  Changlong Gao and
                  Zhangxuan Gu and
                  Yulin Liu and
                  Xinyu Qiu and
                  Shuheng Shen and
                  Yue Wen and
                  Tianyu Xia and
                  Zhenyu Xu and
                  Zhengwen Zeng and
                  Beitong Zhou and
                  Xingran Zhou and
                  Weizhi Chen and
                  Sunhao Dai and
                  Jingya Dou and
                  Yichen Gong and
                  Yuan Guo and
                  Zhenlin Guo and
                  Feng Li and
                  Qian Li and
                  Jinzhen Lin and
                  Yuqi Zhou and
                  Linchao Zhu and
                  Liang Chen and
                  Zhenyu Guo and
                  Changhua Meng and
                  Weiqiang Wang},
  title        = {UI-Venus-1.5 Technical Report},
  journal      = {CoRR},
  volume       = {abs/2602.09082},
  year         = {2026},
  url          = {https://doi.org/10.48550/arXiv.2602.09082},
  doi          = {10.48550/ARXIV.2602.09082},
  eprinttype   = {arXiv},
  eprint       = {2602.09082},
  bibsource    = {dblp computer science bibliography, https://dblp.org}
}

@article{zhang2025mvp,
  author       = {Yunzhu Zhang and
                  Zeyu Pan and
                  Zhengwen Zeng and
                  Shuheng Shen and
                  Changhua Meng and
                  Linchao Zhu},
  title        = {{MVP:} Multiple View Prediction Improves {GUI} Grounding},
  journal      = {CoRR},
  volume       = {abs/2512.08529},
  year         = {2025},
  url          = {https://doi.org/10.48550/arXiv.2512.08529},
  doi          = {10.48550/ARXIV.2512.08529},
  eprinttype   = {arXiv},
  eprint       = {2512.08529},
}

@article{li2026qwen3vlreranker,
  author       = {Mingxin Li and
                  Yanzhao Zhang and
                  Dingkun Long and
                  Keqin Chen and
                  Sibo Song and
                  Shuai Bai and
                  Zhibo Yang and
                  Pengjun Xie and
                  An Yang and
                  Dayiheng Liu and
                  Jingren Zhou and
                  Junyang Lin},
  title        = {Qwen3-VL-Embedding and Qwen3-VL-Reranker: {A} Unified Framework for
                  State-of-the-Art Multimodal Retrieval and Ranking},
  journal      = {CoRR},
  volume       = {abs/2601.04720},
  year         = {2026},
  url          = {https://doi.org/10.48550/arXiv.2601.04720},
  doi          = {10.48550/ARXIV.2601.04720},
  eprinttype   = {arXiv},
  eprint       = {2601.04720},
}

@article{nogueira2019reranking,
  author       = {Rodrigo Nogueira and
                  Kyunghyun Cho},
  title        = {Passage Re-ranking with {BERT}},
  journal      = {CoRR},
  volume       = {abs/1901.04085},
  year         = {2019},
  doi          = {10.48550/arXiv.1901.04085},
  url          = {https://arxiv.org/abs/1901.04085},
  eprinttype   = {arXiv},
  eprint       = {1901.04085},
}

@inproceedings{nayak2025uivision,
  author       = {Shravan Nayak and
                  Xiangru Jian and
                  Kevin Qinghong Lin and
                  Juan A. Rodr{\'{\i}}guez and
                  Montek Kalsi and
                  Nicolas Chapados and
                  M. Tamer {\"{O}}zsu and
                  Aishwarya Agrawal and
                  David V{\'{a}}zquez and
                  Christopher Pal and
                  Perouz Taslakian and
                  Spandana Gella and
                  Sai Rajeswar},
  editor       = {Aarti Singh and
                  Maryam Fazel and
                  Daniel Hsu and
                  Simon Lacoste{-}Julien and
                  Felix Berkenkamp and
                  Tegan Maharaj and
                  Kiri Wagstaff and
                  Jerry Zhu},
  title        = {UI-Vision: {A} Desktop-centric {GUI} Benchmark for Visual Perception
                  and Interaction},
  booktitle    = {Proceedings of the 42nd International Conference on Machine Learning},
  series       = {Proceedings of Machine Learning Research},
  volume       = {267},
  pages        = {45817--45851},
  publisher    = {PMLR},
  address      = {Vancouver, BC, Canada},
  year         = {2025},
  url          = {https://proceedings.mlr.press/v267/nayak25a.html},
}

@article{bai2025qwen3vl,
  author       = {Qwen Team},
  title        = {Qwen3-VL Technical Report},
  journal      = {CoRR},
  volume       = {abs/2511.21631},
  year         = {2025},
  url          = {https://doi.org/10.48550/arXiv.2511.21631},
  doi          = {10.48550/ARXIV.2511.21631},
  eprinttype   = {arXiv},
  eprint       = {2511.21631},
}

@inproceedings{xiao2024hivg,
  author       = {Linhui Xiao and
                  Xiaoshan Yang and
                  Fang Peng and
                  Yaowei Wang and
                  Changsheng Xu},
  editor       = {Jianfei Cai and
                  Mohan S. Kankanhalli and
                  Balakrishnan Prabhakaran and
                  Susanne Boll and
                  Ramanathan Subramanian and
                  Liang Zheng and
                  Vivek K. Singh and
                  Pablo C{\'{e}}sar and
                  Lexing Xie and
                  Dong Xu},
  title        = {HiVG: Hierarchical Multimodal Fine-grained Modulation for Visual Grounding},
  booktitle    = {Proceedings of the 32nd {ACM} International Conference on Multimedia,
                  {MM} 2024, Melbourne, VIC, Australia, 28 October 2024 - 1 November
                  2024},
  pages        = {5460--5469},
  publisher    = {{ACM}},
  year         = {2024},
  doi          = {10.1145/3664647.3681071},
  url          = {https://doi.org/10.1145/3664647.3681071},
}

@inproceedings{yao2024mmca,
  author       = {Ruilin Yao and
                  Shengwu Xiong and
                  Yichen Zhao and
                  Yi Rong},
  editor       = {Jianfei Cai and
                  Mohan S. Kankanhalli and
                  Balakrishnan Prabhakaran and
                  Susanne Boll and
                  Ramanathan Subramanian and
                  Liang Zheng and
                  Vivek K. Singh and
                  Pablo C{\'{e}}sar and
                  Lexing Xie and
                  Dong Xu},
  title        = {Visual Grounding with Multi-modal Conditional Adaptation},
  booktitle    = {Proceedings of the 32nd {ACM} International Conference on Multimedia,
                  {MM} 2024, Melbourne, VIC, Australia, 28 October 2024 - 1 November
                  2024},
  pages        = {3877--3886},
  publisher    = {{ACM}},
  year         = {2024},
  doi          = {10.1145/3664647.3681256},
  url          = {https://doi.org/10.1145/3664647.3681256},
}

@inproceedings{yu2016modeling,
  author       = {Licheng Yu and
                  Patrick Poirson and
                  Shan Yang and
                  Alexander C. Berg and
                  Tamara L. Berg},
  title        = {Modeling Context in Referring Expressions},
  booktitle    = {Computer Vision - {ECCV} 2016 - 14th European Conference, Amsterdam,
                  The Netherlands, October 11-14, 2016, Proceedings, Part {II}},
  series       = {Lecture Notes in Computer Science},
  pages        = {69--85},
  publisher    = {Springer},
  year         = {2016},
  doi          = {10.1007/978-3-319-46475-6_5},
  url          = {https://doi.org/10.1007/978-3-319-46475-6_5},
}

@inproceedings{plummer2015flickr30k,
  author       = {Bryan A. Plummer and
                  Liwei Wang and
                  Chris M. Cervantes and
                  Juan C. Caicedo and
                  Julia Hockenmaier and
                  Svetlana Lazebnik},
  title        = {Flickr30k Entities: Collecting Region-to-Phrase Correspondences for
                  Richer Image-to-Sentence Models},
  booktitle    = {2015 {IEEE} International Conference on Computer Vision, {ICCV} 2015,
                  Santiago, Chile, December 7-13, 2015},
  pages        = {2641--2649},
  publisher    = {{IEEE} Computer Society},
  year         = {2015},
  doi          = {10.1109/ICCV.2015.303},
  url          = {https://doi.org/10.1109/ICCV.2015.303},
}

@inproceedings{mao2016generation,
  author       = {Junhua Mao and
                  Jonathan Huang and
                  Alexander Toshev and
                  Oana Camburu and
                  Alan L. Yuille and
                  Kevin Murphy},
  title        = {Generation and Comprehension of Unambiguous Object Descriptions},
  booktitle    = {2016 {IEEE} Conference on Computer Vision and Pattern Recognition,
                  {CVPR} 2016, Las Vegas, NV, USA, June 27-30, 2016},
  pages        = {11--20},
  publisher    = {{IEEE} Computer Society},
  year         = {2016},
  doi          = {10.1109/CVPR.2016.9},
  url          = {https://doi.org/10.1109/CVPR.2016.9},
}

@article{qiao2021referring,
  author       = {Yanyuan Qiao and
                  Chaorui Deng and
                  Qi Wu},
  title        = {Referring Expression Comprehension: {A} Survey of Methods and Datasets},
  journal      = {{IEEE} Trans. Multim.},
  volume       = {23},
  pages        = {4426--4440},
  year         = {2021},
  doi          = {10.1109/TMM.2020.3042066},
}

@inproceedings{wei2024uniir,
  author       = {Cong Wei and
                  Yang Chen and
                  Haonan Chen and
                  Hexiang Hu and
                  Ge Zhang and
                  Jie Fu and
                  Alan Ritter and
                  Wenhu Chen},
  title        = {UniIR: Training and Benchmarking Universal Multimodal Information
                  Retrievers},
  booktitle    = {Computer Vision - {ECCV} 2024 - 18th European Conference, Milan, Italy,
                  September 29-October 4, 2024, Proceedings, Part {LXXXVII}},
  series       = {Lecture Notes in Computer Science},
  pages        = {387--404},
  publisher    = {Springer},
  year         = {2024},
  doi          = {10.1007/978-3-031-73021-4_23},
  url          = {https://doi.org/10.1007/978-3-031-73021-4_23},
}

@inproceedings{faysse2025colpali,
  author       = {Manuel Faysse and
                  Hugues Sibille and
                  Tony Wu and
                  Bilel Omrani and
                  Gautier Viaud and
                  C{\'{e}}line Hudelot and
                  Pierre Colombo},
  title        = {ColPali: Efficient Document Retrieval with Vision Language Models},
  booktitle    = {The Thirteenth International Conference on Learning Representations,
                  {ICLR} 2025, Singapore, April 24-28, 2025},
  publisher    = {OpenReview.net},
  year         = {2025},
  url          = {https://openreview.net/forum?id=ogjBpZ8uSi},
}

@inproceedings{lin2025mmrerank,
  author       = {Sheng{-}Chieh Lin and
                  Chankyu Lee and
                  Mohammad Shoeybi and
                  Jimmy Lin and
                  Bryan Catanzaro and
                  Wei Ping},
  title        = {Mm-Embed: Universal Multimodal Retrieval with Multimodal {LLMS}},
  booktitle    = {The Thirteenth International Conference on Learning Representations,
                  {ICLR} 2025, Singapore, April 24-28, 2025},
  publisher    = {OpenReview.net},
  year         = {2025},
  url          = {https://openreview.net/forum?id=i45NQb2iKO},
}

@inproceedings{kamath2021mdetr,
  author       = {Aishwarya Kamath and
                  Mannat Singh and
                  Yann LeCun and
                  Gabriel Synnaeve and
                  Ishan Misra and
                  Nicolas Carion},
  title        = {{MDETR} - Modulated Detection for End-to-End Multi-Modal Understanding},
  booktitle    = {2021 {IEEE/CVF} International Conference on Computer Vision, {ICCV}
                  2021, Montreal, QC, Canada, October 10-17, 2021},
  pages        = {1760--1770},
  publisher    = {{IEEE}},
  year         = {2021},
  doi          = {10.1109/ICCV48922.2021.00180},
  url          = {https://doi.org/10.1109/ICCV48922.2021.00180},
}

@inproceedings{li2022glip,
  author       = {Liunian Harold Li and
                  Pengchuan Zhang and
                  Haotian Zhang and
                  Jianwei Yang and
                  Chunyuan Li and
                  Yiwu Zhong and
                  Lijuan Wang and
                  Lu Yuan and
                  Lei Zhang and
                  Jenq{-}Neng Hwang and
                  Kai{-}Wei Chang and
                  Jianfeng Gao},
  title        = {Grounded Language-Image Pre-training},
  booktitle    = {{IEEE/CVF} Conference on Computer Vision and Pattern Recognition,
                  {CVPR} 2022, New Orleans, LA, USA, June 18-24, 2022},
  pages        = {10955--10965},
  publisher    = {{IEEE}},
  year         = {2022},
  doi          = {10.1109/CVPR52688.2022.01069},
  url          = {https://doi.org/10.1109/CVPR52688.2022.01069},
}

@inproceedings{deng2021transvg,
  author       = {Jiajun Deng and
                  Zhengyuan Yang and
                  Tianlang Chen and
                  Wengang Zhou and
                  Houqiang Li},
  title        = {TransVG: End-to-End Visual Grounding with Transformers},
  booktitle    = {2021 {IEEE/CVF} International Conference on Computer Vision, {ICCV}
                  2021, Montreal, QC, Canada, October 10-17, 2021},
  pages        = {1749--1759},
  publisher    = {{IEEE}},
  year         = {2021},
  doi          = {10.1109/ICCV48922.2021.00179},
  url          = {https://doi.org/10.1109/ICCV48922.2021.00179},
}

@inproceedings{yang2022unitab,
  author       = {Zhengyuan Yang and
                  Zhe Gan and
                  Jianfeng Wang and
                  Xiaowei Hu and
                  Faisal Ahmed and
                  Zicheng Liu and
                  Yumao Lu and
                  Lijuan Wang},
  title        = {UniTAB: Unifying Text and Box Outputs for Grounded Vision-Language
                  Modeling},
  booktitle    = {Computer Vision - {ECCV} 2022 - 17th European Conference, Tel Aviv,
                  Israel, October 23-27, 2022, Proceedings, Part {XXXVI}},
  series       = {Lecture Notes in Computer Science},
  pages        = {521--539},
  publisher    = {Springer},
  year         = {2022},
  doi          = {10.1007/978-3-031-20059-5_30},
  url          = {https://doi.org/10.1007/978-3-031-20059-5_30},
}

@article{jiang2025zoomclick,
  author       = {Zhiyuan Jiang and
                  Shenghao Xie and
                  Wenyi Li and
                  Wenqiang Zu and
                  Peihang Li and
                  Jiahao Qiu and
                  Siqi Pei and
                  Lei Ma and
                  Tiejun Huang and
                  Mengdi Wang and
                  Shilong Liu},
  title        = {Zoom in, Click out: Unlocking and Evaluating the Potential of Zooming
                  for {GUI} Grounding},
  journal      = {CoRR},
  volume       = {abs/2512.05941},
  year         = {2025},
  url          = {https://doi.org/10.48550/arXiv.2512.05941},
  doi          = {10.48550/ARXIV.2512.05941},
  eprinttype   = {arXiv},
  eprint       = {2512.05941},
}

@article{li2025cog,
  author       = {Aiden Yiliu Li and
                  Bizhi Yu and
                  Daoan Lei and
                  Tianhe Ren and
                  Shilong Liu},
  title        = {Chain-of-Ground: Improving {GUI} Grounding via Iterative Reasoning
                  and Reference Feedback},
  journal      = {CoRR},
  volume       = {abs/2512.01979},
  year         = {2025},
  url          = {https://doi.org/10.48550/arXiv.2512.01979},
  doi          = {10.48550/ARXIV.2512.01979},
  eprinttype   = {arXiv},
  eprint       = {2512.01979},
}

@article{tang2025focus,
  author       = {Fei Tang and
                  Yongliang Shen and
                  Hang Zhang and
                  Siqi Chen and
                  Guiyang Hou and
                  Wenqi Zhang and
                  Wenqiao Zhang and
                  Kaitao Song and
                  Weiming Lu and
                  Yueting Zhuang},
  title        = {Think Twice, Click Once: Enhancing {GUI} Grounding via Fast and Slow
                  Systems},
  journal      = {CoRR},
  volume       = {abs/2503.06470},
  year         = {2025},
  url          = {https://doi.org/10.48550/arXiv.2503.06470},
  doi          = {10.48550/ARXIV.2503.06470},
  eprinttype   = {arXiv},
  eprint       = {2503.06470},
}

@inproceedings{ye2025guiarp,
  author       = {Xianhang Ye and
                  Yiqing Li and
                  Wei Dai and
                  Miancan Liu and
                  Ziyuan Chen and
                  Zhangye Han and
                  Hongbo Min and
                  Jinkui Ren and
                  Xiantao Zhang and
                  Wen Yang and
                  Zhi Jin},
  title        = {{GUI-ARP:} Enhancing Grounding with Adaptive Region Perception for
                  {GUI} Agents},
  booktitle    = {{ICASSP} 2026 - 2026 {IEEE} International Conference on Acoustics,
                  Speech and Signal Processing},
  pages        = {3676--3680},
  publisher    = {{IEEE}},
  year         = {2026},
  doi          = {10.1109/ICASSP55912.2026.11461773},
  url          = {https://doi.org/10.1109/ICASSP55912.2026.11461773},
}

@misc{lee2025gold,
title={{GOLD}: Global Overview to Local Detail in Efficient Visual Grounding for {GUI} Agents},
author={Mingyu Kim and Jeonghoon Park and Hojun Lee and Taesik Gong},
year={2026},
url={https://openreview.net/forum?id=PVwSDvUWtr}
}

@article{liu2025visurf,
  author       = {Yuqi Liu and
                  Liangyu Chen and
                  Jiazhen Liu and
                  Mingkang Zhu and
                  Zhisheng Zhong and
                  Bei Yu and
                  Jiaya Jia},
  title        = {ViSurf: Visual Supervised-and-Reinforcement Fine-Tuning for Large
                  Vision-and-Language Models},
  journal      = {CoRR},
  volume       = {abs/2510.10606},
  year         = {2025},
  url          = {https://doi.org/10.48550/arXiv.2510.10606},
  doi          = {10.48550/ARXIV.2510.10606},
  eprinttype   = {arXiv},
  eprint       = {2510.10606},
}

@inproceedings{HuSWALWWC22,
  author       = {Edward J. Hu and
                  Yelong Shen and
                  Phillip Wallis and
                  Zeyuan Allen{-}Zhu and
                  Yuanzhi Li and
                  Shean Wang and
                  Lu Wang and
                  Weizhu Chen},
  title        = {LoRA: Low-Rank Adaptation of Large Language Models},
  booktitle    = {The Tenth International Conference on Learning Representations, {ICLR}
                  2022, Virtual Event, April 25-29, 2022},
  publisher    = {OpenReview.net},
  year         = {2022},
  url          = {https://openreview.net/forum?id=nZeVKeeFYf9},
}

@inproceedings{cao2007listwise,
  author       = {Zhe Cao and
                  Tao Qin and
                  Tie{-}Yan Liu and
                  Ming{-}Feng Tsai and
                  Hang Li},
  title        = {Learning to rank: from pairwise approach to listwise approach},
  booktitle    = {Machine Learning, Proceedings of the Twenty-Fourth International Conference
                  {(ICML} 2007), Corvallis, Oregon, USA, June 20-24, 2007},
  series       = {{ACM} International Conference Proceeding Series},
  pages        = {129--136},
  publisher    = {{ACM}},
  year         = {2007},
  doi          = {10.1145/1273496.1273513},
  url          = {https://doi.org/10.1145/1273496.1273513},

}

\end{document}